\documentclass[10pt,twocolumn,letterpaper]{article}

\usepackage{cvpr}              

\usepackage{graphicx}
\usepackage{amsmath}
\usepackage{amssymb}
\usepackage{booktabs}
\usepackage{ulem}
\usepackage{amsfonts}
\usepackage{dsfont}
\usepackage{tikz}
\usepackage{floatrow}
\usepackage{wrapfig}
\usepackage[symbol]{footmisc}
\usepackage[accsupp]{axessibility}  

\usepackage{dblfloatfix}
\usepackage{enumitem}

\usepackage{float}
\floatstyle{plaintop}
\restylefloat{table}
\newfloatcommand{capbtabbox}{table}[][\FBwidth]

\def\checkmark{\tikz\fill[scale=0.4](0,.35) -- (.25,0) -- (1,.7) -- (.25,.15) -- cycle;} 
\newcommand{\zy}[1]{\textcolor{black}{#1}}
\newcommand{\cw}[1]{\textcolor{black}{#1}}

\usepackage{amsmath,amsfonts,bm,mathtools,amsthm}

\def\eqref#1{equation~\ref{#1}}

\def\1{\bm{1}}

\DeclareMathAlphabet{\mathsfit}{\encodingdefault}{\sfdefault}{m}{sl}
\SetMathAlphabet{\mathsfit}{bold}{\encodingdefault}{\sfdefault}{bx}{n}

\newcommand{\R}{\mathbb{R}}

\newcommand{\softmax}{\mathrm{softmax}}

\usepackage[pagebackref,breaklinks,colorlinks]{hyperref}

\usepackage[capitalize]{cleveref}
\crefname{section}{Sec.}{Secs.}
\Crefname{section}{Section}{Sections}
\Crefname{table}{Table}{Tables}
\crefname{table}{Tab.}{Tabs.}

\def\cvprPaperID{12868} 
\def\confName{CVPR}
\def\confYear{2023}

\begin{document}


%

\title{Hint-Aug: Drawing Hints from Foundation Vision Transformers towards Boosted Few-shot Parameter-Efficient Tuning}


\author{Zhongzhi Yu\textsuperscript{1}, \, Shang Wu\textsuperscript{2}, \, Yonggan Fu\textsuperscript{1}, \, Shunyao Zhang\textsuperscript{2}, \, Yingyan (Celine) Lin\textsuperscript{1}\\
\textsuperscript{1}Georgia Institute of Technology \,
\textsuperscript{2}Rice University \\
\small{\{zyu401, yfu314, celine.lin\}@gatech.edu \, \{sw99,sz74\}@rice.edu}}

\maketitle

\vspace{-0.5em}
\begin{abstract}
\vspace{-0.5em}
     Despite the growing demand for tuning foundation vision transformers (FViTs) on downstream tasks, fully unleashing FViTs' potential under data-limited scenarios (e.g., few-shot tuning) remains a challenge due to FViTs' data-hungry nature. Common data augmentation techniques fall short in this context due to the limited features contained in the few-shot tuning data. To tackle this challenge, we first identify an opportunity for FViTs in few-shot tuning: pretrained FViTs themselves have already learned highly representative features from large-scale pretraining data, which are fully preserved during widely used parameter-efficient tuning. We thus hypothesize that leveraging those learned features to augment the tuning data can boost the effectiveness of few-shot FViT tuning. To this end, we propose a framework called \textbf{Hint}-based Data \textbf{Aug}mentation (\textbf{Hint-Aug}), which aims to boost FViT in few-shot tuning by augmenting the over-fitted parts of tuning samples with the learned features of pretrained FViTs. Specifically, Hint-Aug integrates two key enablers: (1) an \textbf{A}ttentive \textbf{O}ver-fitting \textbf{D}etector (\textbf{AOD}) to detect over-confident patches of foundation ViTs for potentially alleviating their over-fitting on the few-shot tuning data and (2) a \textbf{C}onfusion-based \textbf{F}eature \textbf{I}nfusion (\textbf{CFI}) module to infuse easy-to-confuse features from the pretrained FViTs with the over-confident patches detected by the above AOD in order to enhance the feature diversity during tuning. Extensive experiments and ablation studies on five datasets and three parameter-efficient tuning techniques consistently validate Hint-Aug's effectiveness: $0.04\% \sim 32.91\%$ higher accuracy over the state-of-the-art (SOTA) data augmentation method under various low-shot settings. For example, on the Pet dataset, Hint-Aug achieves a 2.22\% higher accuracy with 50\% less training data over SOTA data augmentation methods. 
\end{abstract}

\vspace{-1.2em}
\section{Introduction}
\vspace{-0.3em}
Foundation vision transformers (FViTs)~\cite{dosovitskiy2020image,touvron2021training,touvron2022deit,liu2021swin,wu2021cvt} with billions of floating point operations (FLOPs) and parameters have recently demonstrated significant potential in various downstream tasks~\cite{liu2021swin,liu2021swin2}. The success of FViTs has \cw{ushered in }
a new paradigm in deep learning: pretraining-then-tuning~\cite{yuan2021florence,liu2021swin2,dosovitskiy2020image}, which first pretrains an FViT on a large-scale dataset, then uses recently developed parameter-efficient tuning methods (e.g., visual prompt tuning (VPT)~\cite{jia2022visual}, visual prompting~\cite{bahng2022exploring}, LoRA~\cite{hu2021lora}, and Adapter~\cite{zhang2022neural}) to tune pretrained FViTs on downstream tasks with limited tuning data.
\cw{However, although it is highly desirable, effectively tuning pretrained FViTs for real-world applications, especially under few-shot tuning scenarios, remains a particularly challenging task. }
The reason is that although parameter-efficient tuning methods are dedicatedly designed for FViTs and can alleviate the overfitting issue by reducing the number of trainable parameters~\cite{bahng2022exploring,jia2022visual,zhang2022neural}, the data-hungry nature of FViTs~\cite{dosovitskiy2020image,touvron2021training} is not mitigated and thus the achievable accuracy under data-limited scenarios (e.g., few-shot tuning scenarios) are still limited. 
Therefore, how to effectively tune pretrained FViTs on various downstream tasks with few-shot tuning is still an open question.

\footnotetext[2]{Our code is available at \url{https://github.com/GATECH-EIC/Hint-Aug}}

\begin{figure}[t]
    \centering
    \includegraphics[width=0.85\linewidth]{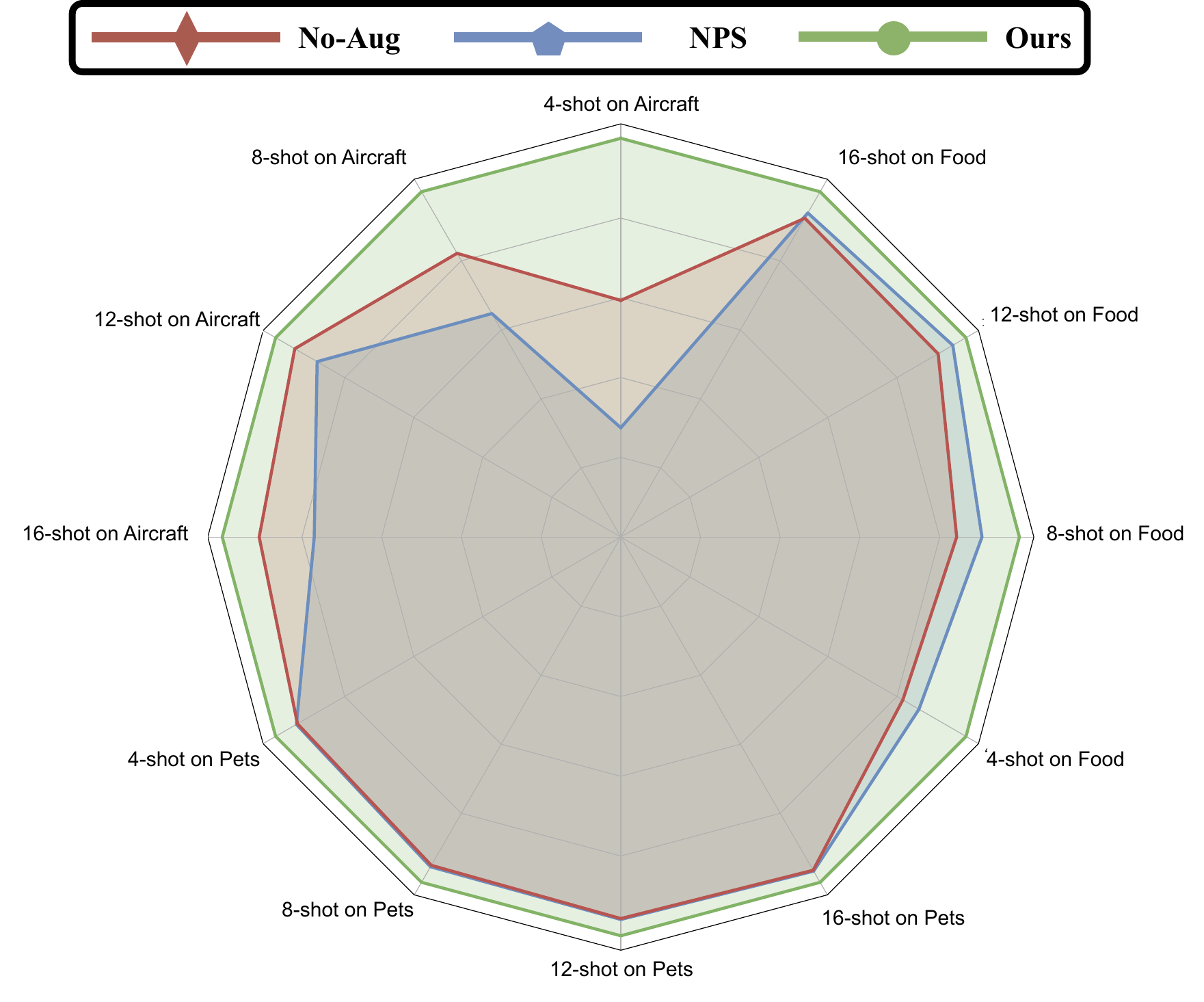}
    \vspace{-1em}
    \caption{The normalized achieved accuracies when few-shot tuning the ViT-base model~\cite{dosovitskiy2020image} on various datasets and numbers of tuning shots using (1) vanilla training without augmentation (i.e., No-Aug), (2) the SOTA parameter-efficient tuning technique~\cite{zhang2022neural} (i.e., NPS), and (3) our proposed Hint-Aug.
    }
    \label{fig:motivation}
\end{figure}

To enhance the effectiveness of parameter-efficient FViT tuning under few-shot settings, one promising direction is to leverage data augmentation techniques to increase the data diversity
and thus the feature diversity of the models when being tuned on few-shot data, boosting the achievable accuracy~\cite{yun2019cutmix,cubuk2018autoaugment,zhang2017mixup,hendrycks2019augmix}. 
Nevertheless, it has been shown that existing data augmentation techniques fall short in boosting the model accuracy under few-shot tuning scenarios. This is because most of the existing data augmentation techniques are random-based (e.g., RandAugment~\cite{cubuk2020randaugment}, AutoAugment~\cite{cubuk2018autoaugment}, color jitter, mixup~\cite{zhang2017mixup}, and cutmix~\cite{yun2019cutmix})\cw{, which }
only randomly permute existing features in the training data and thus cannot generate new and meaningful features~\cite{wang2018low}. As illustrated in Fig.~\ref{fig:motivation}, we observe that neither the widely-used random-based data augmentation techniques (i.e., a dedicated combination of techniques including RandAugment~\cite{cubuk2020randaugment}, color jitter, and random erasing~\cite{zhong2020random} as in~\cite{zhang2022neural}) nor training without data augmentation can consistently achieve a satisfactory accuracy across different datasets under few-shot tuning. Specifically, when being applied to fine-grained classification tasks, e.g., the Aircraft dataset~\cite{maji2013fine}, these random-based data augmentation techniques actually hurt the achievable accuracy. The reason is that random-based data augmentation techniques can easily create out-of-manifold samples~\cite{zhang2017mixup,yun2019cutmix}, especially on commonly used fine-grained datasets. Such out-of-manifold samples can largely degrade the achievable accuracy given the limited number of training samples under few-shot tuning scenarios~\cite{guo2019mixup}. Therefore, it is \cw{crucial }
to develop data augmentation techniques that can adaptively augment the given training samples with diverse\cw{, but still within-manifold, }
features to boost the effectiveness of tuning FViTs on various downstream tasks.

This work sets out to close the increasing gap between the growing demand for effective few-shot FViT tuning and the unsatisfactory achievable accuracy by existing techniques. In particular, we identify that in few-shot parameter-efficient tuning, the pretrained FViTs' weights are fixed during tuning. Meanwhile, existing works have shown that (1) pretrained transformer models have already learned complex but generalizable features~\cite{dosovitskiy2020image,liu2021swin,touvron2021training} and (2) gradient-based methods can extract the learned features from pretrained models and then add them to the input images~\cite{mordvintsev2015inceptionism,madry2017towards}. Therefore, we hypothesize that FViTs' few-shot tuning accuracies can be non-trivially improved by leveraging the learned features in the pretrained FViTs. Specifically, we make the following contributions:

\begin{itemize}[leftmargin=0.5em,topsep=0.em]
    \item We propose a framework called \textbf{Hint}-based Data \textbf{Aug}mentation (\textbf{Hint-Aug}), which is dedicated to boosting the achievable accuracy of FViTs under few-shot tuning scenarios by leveraging the learned features of pretrained FViTs to guide the data augmentation strategy used for the training dataset in an input-adaptive manner.

   \item Our Hint-Aug framework integrates two enablers: \uline{(1)} an \textbf{A}ttentive \textbf{O}ver-fitting \textbf{D}etector (AOD) to identify the over-fitting samples and patches in the given training dataset by making use of the attention maps of pretrained FViTs and \uline{(2)} a \textbf{C}onfusion-based \textbf{F}eature \textbf{I}nfusion (CFI) module to adaptively infuse pretrained FViTs' learned features into the training data to better tuning those models on downstream tasks, alleviating the commonly recognized challenge of having limited features under few-shot tuning. 

     \item Extensive experiments and ablation studies on five datasets and three parameter-efficient tuning techniques consistently validate the effectiveness of our proposed Hint-Aug framework, which achieves a $0.04\%\sim 32.91\%$ higher accuracy over state-of-the-art (SOTA) data augmentation methods~\cite{zhang2022neural} across different datasets and few-shot settings. 
    For example, on the Pets dataset, Hint-Aug achieves a 2.22\% higher accuracy with 50\% less training data compared with the SOTA augmentation method. 
\end{itemize}

\section{Related Works}

\subsection{FViTs}
\label{sec:fvits}
Inspired by the recent success of vision transformers (ViTs), one of the most notable directions in ViTs is to scale up ViTs' model size to build FViTs, aiming to replicate the success of large-scale neural language processing foundation models~\cite{raffel2020exploring,devlin2018bert,floridi2020gpt} in the field of computer vision~\cite{liu2021swin2,dosovitskiy2020image,yuan2021florence}. Existing efforts in developing FViTs mainly fall into two categories: (1) exploring how to scale up ViTs' architectures to construct powerful FViTs~\cite{liu2021swin2,zhai2022scaling,zhou2021deepvit}; (2) developing self-supervised pretraining techniques to train FViTs so that their learned representations can be more effectively generalized to downstream tasks~\cite{he2021masked,feichtenhofer2022masked,caron2021emerging,bao2021beit,li2021efficient}.

Unlike conventional convolutional neural networks (CNNs), FViTs extensively use the self-attention mechanism to extract global features, resulting in improved task accuracy with larger models (e.g., over 10G FLOPs). Specifically, in ViTs, a series of $N$ input image patches $X=[x_1, \cdots, x_N]\cw{^\top}\in \R^{N\times D}$, where $D$ is the embedding dimension, is sequentially processed by ViT blocks. In each block, the input is first converted into queries $Q\in \R^{N\times d}$, keys $K\in \R^{N\times d}$ and values $V\in \R^{N\times d}$ ($d$ denotes the hidden dimension) via linear projection, followed by the computation of the self-attention, which is calculated as:

 \begin{equation}
    \text{Attention}(Q,K,V) = \softmax(\frac{QK^T}{\sqrt{d}})V
\end{equation}

The outputs are then fed into a feed-forward network to extract information in the channel dimension.

\subsection{Parameter-efficient Tuning}
\label{sec:parameter-tuning}
Motivated by the impressive pretraining performance of FViTs on large-scale datasets, there has been a growing interest in applying FViTs to real-world applications. The common solution follows the pretraining-then-tuning paradigm, which tunes pretrained FViTs on various downstream tasks based on the corresponding applications' needs. However, with conventional weight tuning, each task would need to store an additional set of model weights, which can lead to cumbersome and prohibitive storage overhead. To this end, various parameter-efficient tuning methods have been proposed~\cite{hu2021lora,zhang2022neural,bahng2022exploring,sung2022lst}. 
In parameter-efficient tuning, a set of tiny learnable modules are added to the pretrained FViTs, while the weights of the backbone FViTs remain unchanged during tuning~\cite{hu2021lora,jia2022visual,houlsby2019parameter}. This approach offers two benefits:
\uline{(1)} it allows FViTs to be tuned on new downstream tasks with negligible additional parameters, and \uline{(2)} the pretrained FViTs can be easily retrieved at any time by simply removing the added parameter-efficient tuning modules.

Among recent parameter-efficient tuning techniques, LoRA~\cite{hu2021lora} proposes to learn a set of low-rank weights and apply them on top of the backbone weights, and VPT~\cite{jia2022visual} proposes to use the idea of prompt tuning, inserting a set of task-specific prompts as additional tokens. More recently, NPS~\cite{zhang2022neural} proposes to search for the optimal combination of parameter-efficient tuning techniques and their corresponding hyperparameters through neural architecture search.

\subsection{Few-shot Tuning}
\label{sec:fewshot}
Few-shot tuning aims to tune pretrained models on new tasks with limited samples per class~\cite{long2015learning,liu2018learning,guo2020broader,frikha2021few,finn2017model}. It has gained increasing attention in recent years~\cite{volpi2018generalizing} as high-quality data is scarce in many real-world applications~\cite{bansal2022systematic}. Recently, a few pioneering works that target few-shot tuning for ViTs propose to customize meta-learning tasks and learning objectives under the guidance of self-attention modules~\cite{chen2022mask,liu2020universal,chen2021sparse,wang2022few,xu2023exploring}. In this paper, we aim to enhance FViTs' few-shot tuning accuracy from an orthogonal direction, i.e., adaptively augmenting the few-shot tuning samples to compensate for their lack of diverse features.

\subsection{Data Augmentation}
Data augmentation aims to enhance data diversity and thus the feature diversity of the models~\cite{zhang2017mixup,yun2019cutmix,hendrycks2019augmix,wang2021augmax,cheng2021deepmix,gong2021keepaugment,zhong2020random,chen2021transmix,qin2021understanding}. An effective data augmentation strategy should properly enhance data diversity, while simultaneously avoiding the generation of out-of-manifold data caused by excessive augmentation intensity~\cite{zhang2017mixup,venkataramanan2021alignmix}. 
Although various data augmentation techniques have been proposed, how to effectively augment the data under few-shot tuning settings is still an open question. The limited data diversity in few-shot data calls for techniques that can generate novel but meaningful features~\cite{wang2018low,wang2020generalizing}. To this end, most existing few-shot data augmentation techniques adopt generative models to generate in-domain data, which, however, further increase the memory and storage overhead of tuning FViTs~\cite{hariharan2017low,li2020adversarial,luo2021few,gao2018low}.

One potential way to alleviate the aforementioned challenges is to use adversarial techniques to generate samples with beneficial features~\cite{volpi2018generalizing,rusak2020simple,ford2019adversarial}. However, the majority of these works focus on improving adversarial robustness instead of the clean accuracy~\cite{rusak2020simple,ford2019adversarial,volpi2018generalizing,zhao2020maximum,zhang2019you,gui2019model}. In contrast, our work explores the opportunities of leveraging adversarial training to generate beneficial features that can boost the clean accuracy during few-shot parameter-efficient tuning. 

\begin{figure*}[t]
    \centering
    \includegraphics[width=0.9\textwidth]{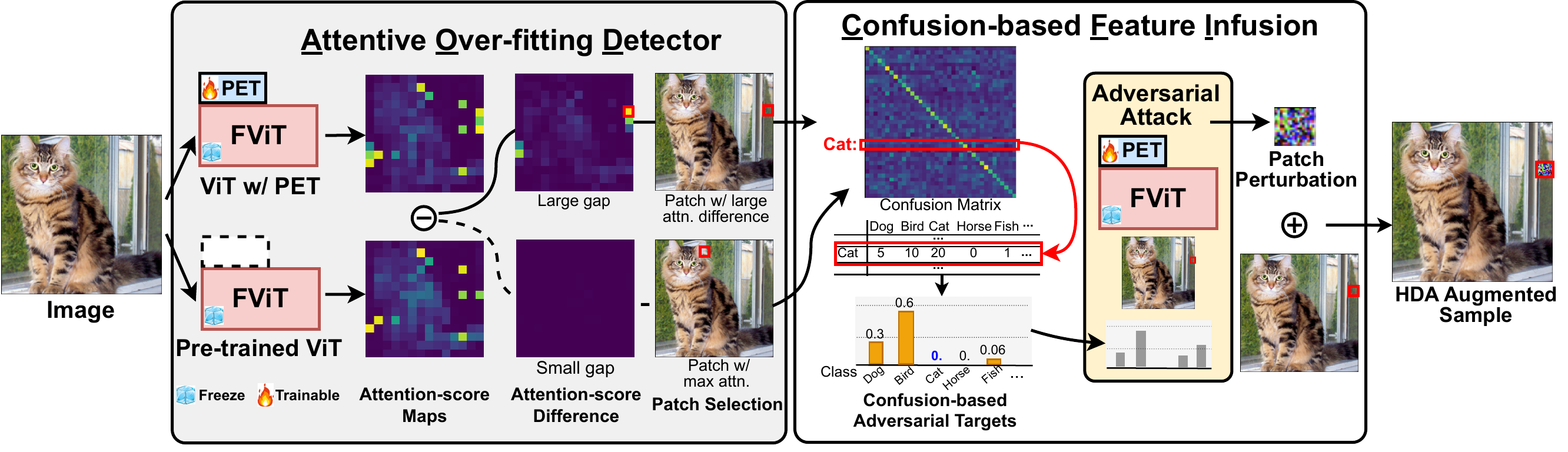}
    \caption{An overview of our proposed Hint-Aug framework, which consists of two enablers: (1) an AOD to detect whether the current sample is prone to over-fitting and which patch is prone to over-fitting, and (2) a CFI to infuse easy-to-confuse features to the over-fitted patches detected by the aformentioned AOD to increase the feature diversity of the tuning data and thus alleviate the over-fitting issue. In the figure, PET represents the parameter-efficient tuning module (e.g., Adapter~\cite{houlsby2019parameter}, VPT~\cite{jia2022visual}, and LoRA~\cite{hu2021lora}) added on top of the pretrained FViTs.}
    \label{fig:framework}
\end{figure*}

\section{The Proposed Hint-Aug Framework}

\subsection{Hint-Aug: Motivation}
\label{sec:motivation}

We first identify that the characteristics of parameter-efficient tuning together with pretrained FViTs provide a unique opportunity for FViTs' parameter-efficient tuning. Based on this observation, 
we then propose our Hint-Aug framework, which utilizes these characteristics to enhance the tuning effectiveness. We describe each of the characteristics in detail below:

\textbf{Characteristics of parameter-efficient tuning}: As mentioned in Sec.~\ref{sec:fvits} and Sec.~\ref{sec:parameter-tuning}, the weights of pretrained FViTs are fixed during tuning. Therefore, the tuned FViTs behave the same as their pretrained counterpart after the added tuning modules (e.g., those adopted in Adapter~\cite{houlsby2019parameter}, VPT~\cite{jia2022visual}, and LoRA~\cite{hu2021lora}) are removed~\cite{zhang2022neural}. This motivates us to consider whether we can make use of this characteristic to improve the achievable few-shot tuning accuracy by leveraging the pretrained FViTs.

\textbf{Characteristics of pretrained FViTs}: Existing works have shown that pretrained FViTs have two promising characteristics regarding their learned features: (1) pretrained FViTs can identify complex but meaningful features~\cite{dosovitskiy2020image,liu2021swin}, even on unseen datasets without tuning~\cite{li2021efficient,he2021masked,cao2022understand}; 
(2) the learned features in FViTs can be reversely projected to the input image space using gradient-based methods~\cite{madry2017towards,fu2022patch,mordvintsev2015inceptionism}.

Given the aforementioned characteristics of both parameter-efficient tuning and pretrained FViTs, we hypothesize that these characteristics provide a unique opportunity to effectively leverage the pretrained FViTs to augment the few-shot tuning data. To validate our hypothesis, we aim to explore proper ways to leverage the learned features in pretrained FViTs to boost the effectiveness of few-shot tuning. Specifically, given the two commonly recognized major challenges of few-shot tuning, which are over-fitting~\cite{antoniou2017data,vinyals2016matching} and the lack of data diversity in the tuning data~\cite{wang2018low,wang2020generalizing}, we set out to answer the following questions: \textbf{Q1} - \textit{Can pretrained FViTs detect the potential over-fitting issue during few-shot tuning?} and \textbf{Q2} - \textit{Can we leverage pretrained FViTs to generate meaningful features for enhancing the data diversity?} Our proposed Hint-Aug framework provides an effective solution to these two questions.

\subsection{Hint-Aug: Overview}
We first give an overview of our proposed Hint-Aug framework, which is dedicatedly designed for few-shot parameter-efficient tuning of FViTs by leveraging the characteristic of parameter-efficient tuning that pretrained FViTs' weights are not updated during tuning, allowing the features learned in pretrained FViTs to be utilized to augment the tuning data. As shown in Fig.~\ref{fig:framework}, Hint-Aug adopts a two-stage detect-then-augment pipeline. In particular,
to answer the above \textbf{Q1}, Hint-Aug uses AOD to detect (1) whether the tuned FViT is over-fitted on this image and (2) which patch in the image is more prone to be over-fitted; To address \textbf{Q2}, Hint-Aug further augments the patch detected from AOD by infusing the easy-to-confuse features with our proposed CFI module.
We introduce our AOD and CFI modules in Sec.~\ref{sec:aod} and Sec.~\ref{sec:cfi}, respectively.

\subsection{Enabler 1: Attentive Over-fitting Detector}
\label{sec:aod}
Over-fitting is a well-known issue in few-shot tuning scenarios \cite{li2020adversarial,wang2020generalizing}
, and becomes even more severe due to the combination of larger model size and limited data size during few-shot FViT tuning. Therefore, our AOD aims to explore whether we can detect the underlying over-fitting issue for each tuning sample on-the-fly during parameter-efficient tuning of FViTs.

Inspired by the various visualizations showing FViTs' attention distributions in previous works~\cite{he2021masked,yu2022mia,steiner2021train,fu2022patch}, we hypothesize that the evolution of attention distributions during the tuning process contains hidden traces for identifying the existence of over-fitting. 
To validate this hypothesis, we utilize an attention-score map to quantify the impact of each input image patch on the FViT's attention distribution.
Specifically, an attention-score map is constructed with the attention-score corresponding to each patch of the input image and we define the attention-score as follows: given the attention distribution $[a_1^{(l,h,i)},\cdots,a_N^{(l,h,i)}]$ for the $i$-th patch of the $h$-th head in the $l$-th layer, the attention-score $s_j^{(l,k)}$ of the $j$-th patch for the $k$-th query patch is defined as:

\begin{equation}
    s_j^{(l,k)} = \sum_{h}a_j^{(l,h,k)}
\end{equation}
For the sake of simplicity, we omit the superscript $l$ and $k$ in the following text.

\begin{figure}[t]
\begin{minipage}{1\textwidth}
    \centering
    \includegraphics[width=0.7\linewidth]{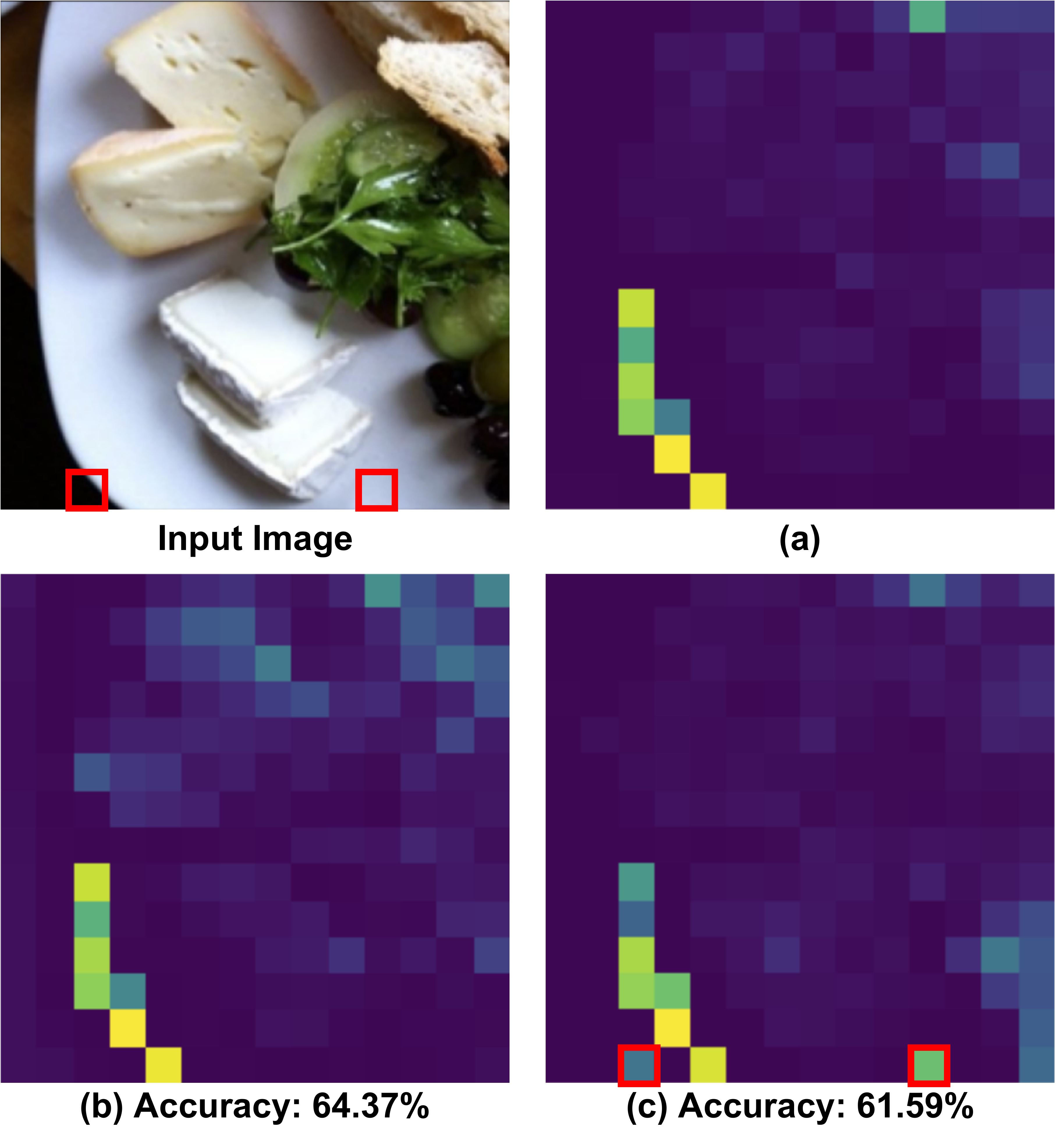}
    \vspace{-0.5em}
    \caption{Visualization of the attention-score map from the (a) pretrained foundation model, called ViT-base, (b) parameter-efficient tuned ViT-base model with 20\% of the total tuning epochs, achieving an accuracy of 64.37\%, and (c) parameter-efficient tuned ViT-base model with an accuracy of 61.55\%.}

    \label{fig:attn_vis}
\end{minipage}
\end{figure}

By visualizing the attention-score at different stages of tuning, as shown in Fig.~\ref{fig:attn_vis}, we can draw two observations: (1) the attention-score map of the pretrained FViT itself (see Fig.~\ref{fig:attn_vis}(a)) shares a high correlation with that of the half-tuned FViT model (see Fig.~\ref{fig:attn_vis}(b)), and a relatively higher tuning accuracy (e.g., 64.37\%) suggests that the over-fitting issue is not severe at the corresponding tuning stage; (2) the attention-score map at the end of tuning (see Fig.~\ref{fig:attn_vis}(c)) focuses 
more on certain patches (marked in \textcolor{red}{red}) that are not focused on by the pretrained FViT, and a lower tuning accuracy (e.g., 61.55\%) indicates the existence of the over-fitting issue. Additionally, we observe that patches with a newly attracted higher attention-score (marked in \textcolor{red}{red}) do not contain human-readable information for identification. 
For example, some patches only consist of a black background that does not contribute to identifying the target class, e.g., a cheese plate. This finding suggests that these patches could be the reason for over-fitting.

Based on the observations above, we propose an AOD module to use the attention-score map difference between the pretrained FViT and the corresponding one being tuned to identify both the existence of over-fitting and which patch contributes most to the over-fitting issue. Specifically, given the attention-score maps $S^P=[s_1^{P}, \cdots,s_N^{P}]$ generated from the pretrained FViT (denoted as $P$) and $S^T=[s_1^{T}, \cdots,s_N^{T}]$ generated from the FViT model to be tuned (denoted as $T$), we define the over-fitting indicator as:

\begin{equation}
I=\left\{
\begin{array}{rl}
0, & \sum_{i}\|s_i^P - s_i^T\| < \lambda\sum_{i}\|s_i^P\| \\
1, & \text{otherwise}
\end{array}
\right.
\label{eq:overfit_indicator}
\end{equation}

\noindent where $\lambda$ is a hyperparameter to control the sensitivity of over-fitting detection. 


When over-fitting occurs (i.e., $I=1$), we select the patch that significantly 
changes the attention-score map as the target patch to be augmented in order to alleviate the over-fitting issue. Thus, we select the patch $p$ to augment, where $p$ is defined by: 
\begin{equation}
    p = \arg\max_i (\|s_i^P - s_i^T \|)
\end{equation}

Otherwise, when there is no detected over-fitting issue, we select the patch $p$ with the highest attention-score as the target patch to be augmented from all patches in the corresponding image.

\subsection{Enabler 2: Confusion-based Feature Infusion}
\label{sec:cfi}
With the selected over-fitted patch detected by the AOD above, the remaining question is how to augment the selected patch with meaningful features to (1) alleviate the over-fitting issue and (2) increase the diversity of tuning data with meaningful features. Therefore, we propose CFI that uses adversarial attack-based methods to extract the learned features from the pretrained FViT model and infuse them into the selected patch with the aim of improving the feature diversity in a meaningful way, thus alleviating the over-fitting issue.

However, achieving a meaningful feature extraction and infusion that can help boost the few-shot tuning accuracy is non-trivial. Naively augmenting samples with commonly used attack objectives (e.g., perturbing the image to reduce the value of the model's output logit on the correct class) can easily lead to out-of-manifold samples, as shown in our alation study in Sec.~\ref{sec:obj}. To overcome this, 
the CFI module incorporates injected features to steer the model prediction towards a synthetic target label. This target label is determined by utilizing a confusion matrix, which quantifies the degree to which the model is prone to confusion between pairs of classes.

Specifically, we construct a confusion matrix $C\in \R^{M\times M}_{\geq0}$ in CFI, where $M$ is the total number of classes. As shown in a recent study on open set detection~\cite{vaze2021open}, a pre-softmax model output has a better ability to preserve a model's uncertainty of samples. We thus define $C$ as follows:
\vspace{-0.6em}
\begin{equation}
    C_{i,j} = \sum_{X:y(X)=j}\left(f_i(X)-\min_{i'}f_{i'}(X)\right)
\end{equation}
\vspace{-1.5em}

\noindent where $i$ and $j$ are coordinates in $C$ that represent two classes; $y$ and $f\in\R^M$ are the ground truth label and pre-softmax output given the input image $X$. The generated confusion matrix $C$ helps to identify the class-wise similarity learned by the model and distinguish the class pairs that are easy to be confused by the model. 

To infuse the easy-to-confuse features to the patch, given input $X$ with label $y$, we propose to design the attack label $\tilde{f}(X)\in\R^M_{\geq0}$ \cw{where the $i$-th element is computed} as:

\begin{equation}
\tilde{f}_i(X) = \left\{
\begin{array}{lr}
\frac{C_{i,y}}{\sum_{j}C_{j, y}-C_{y,y}}, & i\neq y \\
0, & i=y
\end{array}
\right.
\end{equation}

The loss function is defined as 
\begin{equation}
    \mathcal{L}_{tar} = \text{CrossEntropy}(\softmax(f), \softmax(\tilde{f}))
\end{equation}

By optimizing the patch to minimize the above loss, the generated features are further shifted \cw{towards} the direction where the model considers an easy-to-confuse class from the current class. \cw{This shift allows the model to learn to differentiate between the current class and the easy-to-confuse class, effectively extending the decision boundary of the current class.}

\section{Experimental Results}
\subsection{Experimental Setup}
\textbf{Datasets, few-shot settings, models, and parameter-efficient tuning techniques.}
\uline{Datasets and few-shot settings.} We adopt five commonly-used datasets for few-shot tuning, including Food~\cite{bossard2014food}, Pet~\cite{parkhi2012cats}, Cars~\cite{krause20133d}, Flowers~\cite{jia2022visual}, and Aircraft~\cite{maji2013fine}, and benchmark our Hint-Aug under 1/2/4/8/12/16-shot scenarios to provide a thorough evaluation of its achieved accuracy across different few-shot tuning scenarios. 
\uline{Models.} We conduct our experiment on a widely used FViT model, i.e., ViT-Base~\cite{dosovitskiy2020image}. 
\uline{Adopted parameter-efficient tuning methods.} We consider three most widely used parameter-efficient tuning methods including Adapter~\cite{houlsby2019parameter}, LoRA~\cite{hu2021lora}, and VPT~\cite{jia2022visual}.

\textbf{Baselines.}
We benchmark our proposed Hint-Aug against two baselines, including the SOTA data augmentation technique for parameter-efficient FViT tuning introduced in~\cite{zhang2022neural} (denoted as NPS) and the vanilla tuning without augmentation (denoted as No-Aug). It is worth noting that, given the unique challenge of limited data diversity in the few-shot tuning scenarios, even the SOTA data augmentation technique, i.e., the aforementioned NPS~\cite{zhang2022neural}, can lead to an inferior accuracy than that of the vanilla tuning without augmentation (as shown in Fig.~\ref{fig:motivation}). Thus, it is necessary to include No-Aug as one of the baselines.

\textbf{Tuning settings.}
In our experiments, we set $l=5$ and adopt the center patch in each image as the query patch (i.e., $k=90$), following~\cite{fu2022patch}. 
We follow the widely adopted few-shot tuning settings in~\cite{zhang2022neural}. Specifically, we tune the model for 100 epochs using a batch size of 256, a learning rate of 0.01, and an SGD optimizer starting from the ImageNet~\cite{deng2009imagenet} pretrained ViT-Base~\cite{dosovitskiy2020image}. Following NPS~\cite{zhang2022neural}, we also use data augmentation techniques including color-jitter with a factor of 0.4 and RandAugment~\cite{cubuk2020randaugment} with a magnitude of 9 and a standard deviation equal to 0.5. We set $\lambda$ in Eq.~\ref{eq:overfit_indicator} as 0.1 and use FGSM~\cite{goodfellow2014explaining} to generate the adversarial samples with attack radius $\epsilon=0.001$. Additionally, we run all experiments in the paper three times and report the average accuracy, following NPS~\cite{zhang2022neural}.

\subsection{Benchmark on Few-shot Image Classification}
\label{sec:main exp}

\begin{figure*}[t]
    \centering
    \includegraphics[width=0.87\textwidth]{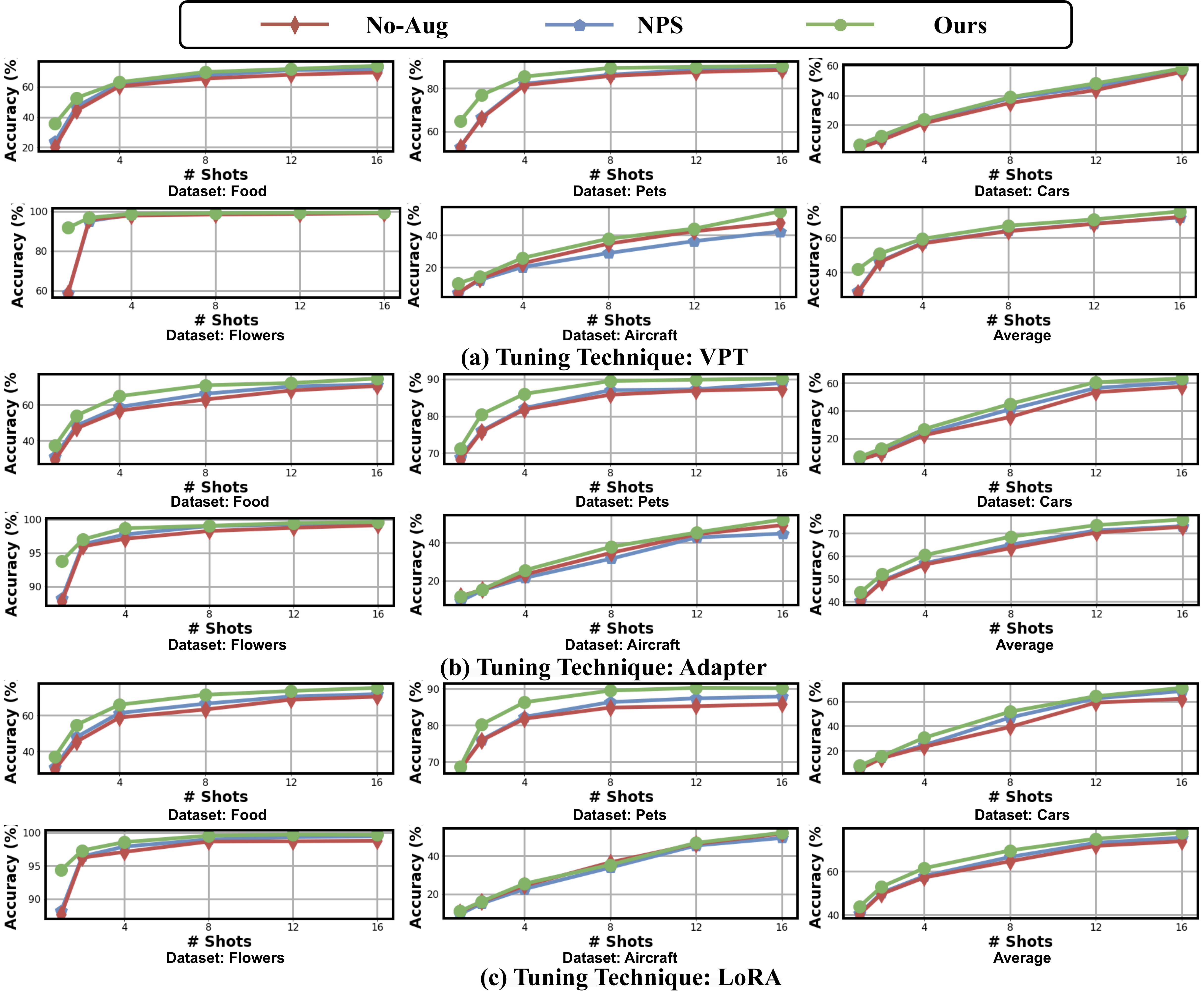}
    \vspace{-0.5em}
    \caption{Benchmark Hint-Aug on Food~\cite{bossard2014food}, Pets~\cite{parkhi2012cats}, Cars~\cite{krause20133d}, Flowers~\cite{nilsback2006visual}, and Aircraft~\cite{maji2013fine} with parameter-efficient tuning methods (a) VPT, (b) Adapter, and (c) LoRA on 1/2/4/8/12/16-shot setting. \vspace{-0.5em}}
    \label{fig:main exp}
\end{figure*}

We first benchmark our proposed method on five commonly used few-shot image classification datasets~\cite{bossard2014food,parkhi2012cats,krause20133d,jia2022visual,maji2013fine} with different parameter-efficient tuning techniques and few-shot settings. As shown in Fig.~\ref{fig:main exp}, although the SOTA augmentation baseline NPS~\cite{zhang2022neural} suffers from considerable accuracy degradation compared with the vanilla tuning method No-Aug on fine-grained image classification dataset (e.g., a 5.55\% accuracy drop on~\cite{maji2013fine}), our proposed Hint-Aug achieves $0.25\%\sim 6.10\%$, $0.10\% \sim 32.91\%$, and $0.04\%\sim 6.17\%$ higher accuracies across different shot selections over baselines when using Adapter~\cite{houlsby2019parameter}, VPT~\cite{jia2022visual}, and LoRA~\cite{hu2021lora} tuning, respectively.

In particular, we draw the following two exciting observations: (1) the features generated by Hint-Aug can compensate for the lack of sufficient tuning data and improve accuracy under more stringent few-shot settings. Specifically, Hint-Aug boosts the accuracy of 8-shot tuning by $2.45\% \sim 4.96\%$ and surpasses the 12-shot tuning with NPS~\cite{zhang2022neural} by a $0.73\% \sim 2.22\%$ higher accuracy when tuning Adapter and LoRA on the Food and Pets datasets; (2) Hint-Aug's ability to extract features from the pretrained FViTs and infuse them into the tuning data can considerably boost accuracy in extreme few-shot scenarios (e.g., 1-shot tuning). For example, on the Pets dataset, tuning VPT with Hint-Aug under a 1-shot setting leads to a 32.91\% higher accuracy than that of NPS~\cite{zhang2022neural}.

\begin{table}[t]
      \caption{Ablation study on each enabler's contribution to the final accuracy.}
      \vspace{-1em}
          \centering
    \resizebox{0.65\linewidth}{!}{
    \begin{tabular}{cc|ccc}
    \toprule
        AOD & CFI & Food & Pets & Cars\\
        \midrule
        &  & 66.25 & 86.97 & 40.83\\
        \checkmark &  & 68.53 & 88.01 & 42.17\\
        & \checkmark & 70.52 & 89.07 & 43.55\\
        \checkmark & \checkmark & 71.04 & 89.42 & 44.80\\
        \bottomrule
    \end{tabular}
    }
    \vspace{-1em}
    \label{tab:breakdown}
\end{table}

\vspace{-0.2em}
\subsection{Ablation Studies}
\vspace{-0.2em}

\subsubsection{Accuracy Improvement Breakdown}

\textbf{Setup.}
To better understand the contribution of each enabler of Hint-Aug, including AOD and CFI, to the final accuracy, we conduct an ablation study where we run 8-shot tuning with Adapter~\cite{houlsby2019parameter} on three datasets, namely Food~\cite{bossard2014food}, Pets~\cite{parkhi2012cats}, and Cars~\cite{krause20133d}. We implement this accuracy improvement breakdown experiment as follows: (1) when using AOD only, we adopt the data augmentation method in~\cite{zhang2022neural} to augment the selected patch; (2) when using CFI only, we generate the samples with $\mathcal{L}_{tar}$ loss and randomly select a patch to augment in each image. 

\textbf{Observations.} As shown in Tab.~\ref{tab:breakdown}, when augmenting a selected patch, we can observe that (1) enabling either AOD or CFI can lead to an accuracy improvement of $1.04\% \sim 2.28\%$ and $2.10\% \sim 4.27\%$ over the baseline (e.g., neither AOD nor CFI enabled), respectively. This indicates that both key challenges (i.e., the over-fitting issue and lack of feature diversity as analyzed in Sec.~\ref{sec:motivation}) indeed hurt the achievable accuracy of few-shot tuning and our proposed enablers can effectively alleviate the challenge in over-fitting; (2) Combining both AOD and CFI can marry the merit of both, thus further boosting the achievable accuracy by $0.35\% \sim 2.63\%$ over that of enabling only one of AOD or CFI.

\vspace{-0.6em}
\subsubsection{Ablation on Adversarial Objectives}
\label{sec:obj}

\textbf{Setup.}
We conduct ablation studies to validate the choice of loss functions for generating the adversarial sample for feature infusion. As mentioned in Sec.~\ref{sec:cfi} and Sec.~\ref{sec:fewshot}, different loss functions can have different impacts on the tuning accuracy and an improper loss function can lead to inferior clean accuracy. In Tab.~\ref{tab:loss func}, we validate the objective function we selected with other potential candidates when tuning on the Food dataset~\cite{bossard2014food} using Adapter~\cite{houlsby2019parameter}, where ``Full" indicates generating adversarial samples with the whole image, instead of the selected patch, ``Untarget" means using the conventional attack target that minimizes the value of the model's output logit on the correct class by augmenting the selected patch, and ``Random" means augmenting the selected patch to mislead the output of the augmented image toward another randomly selected class.

\begin{table}[t]
    \centering
    \caption{Ablation study on different adversarial objectives. }
    \vspace{-1em}
    \resizebox{0.85\linewidth}{!}{
    \begin{tabular}{c|ccccc}
    \toprule
        Target & Full & Untarget & Random & Proposed\\
        \midrule
        4-shot & 57.49 & 59.21 & 62.35 &64.92 \\ 
        8-shot & 66.04 & 67.36 & 69.14 & 71.04 \\ 
        16-shot & 70.78 & 71.58 & 73.85 & 74.90 \\ 
        \bottomrule
    \end{tabular}
    }
    \vspace{-1em}
    \label{tab:loss func}
\end{table}

\textbf{Observations.}
As shown in Tab.~\ref{tab:loss func}, ``Full" leads to the worst achieved accuracy which is $0.80\%\sim 1.72\%$ lower than the second worst object ``Untarget". ``Untarget" also leads to a $3.32\%\sim5.71\%$ lower accuracy than our proposed method. These two observations suggest that (1) attacking the image as a whole cannot effectively help with FViT tuning, and (2) naively using the ``Untarget" attack can easily lead to out-of-manifold data. \underline{Furthermore}, the $1.78\% \sim 3.14\%$ accuracy improvement of ``Random" over ``Untarget" suggests that despite the simple method of selecting the direction to add features, adding features from other classes can help with tuning. However, the lack of a more precise augmentation direction still limits the achievable accuracy when using the ``Random" adversarial objective, leading to a $1.05\% \sim 2.57\%$ lower accuracy than the adversarial objective adopted in Hint-Aug. 

\vspace{-1em}
\subsubsection{Sensitivity to Augmentation Intensity}
\cw{According to recent studies~\cite{touvron2021training,steiner2021train}, augmentation intensity is a crucial factor in FViT tuning. Thus, we investigate the impact of the adversarial attack radius $\epsilon$ on the achievable accuracy of Hint-Aug.} When tuning with Adapter~\cite{houlsby2019parameter} on Food~\cite{bossard2014food} under an 8-shot setting, Hint-Aug achieves relatively stable achieved accuracy under the drastic change in attack radius. Specifically, as shown in Tab.~\ref{tab:aug_intensity}, increasing or decreasing $\epsilon$ by 5 times only leads to a $0.03\% \sim 0.21\%$ accuracy change, while changing $\epsilon$ by 10 times leads to a $0.89\%\sim 1.03\%$ accuracy change compared with a radius of 0.001 that we select in Hint-Aug, proving the robustness of Hint-Aug in different selections of hyperparameters. It is worth noting that changing $\epsilon$ by 10 times is a non-trivial change. As suggested in~\cite{fu2022patch}, changing $\epsilon$ by 8 times leads to an accuracy change larger than $27.94\%$ when attacking DeiT-Tiny on ImageNet.

\begin{figure}[t]
    \centering
    \includegraphics[width=\linewidth]{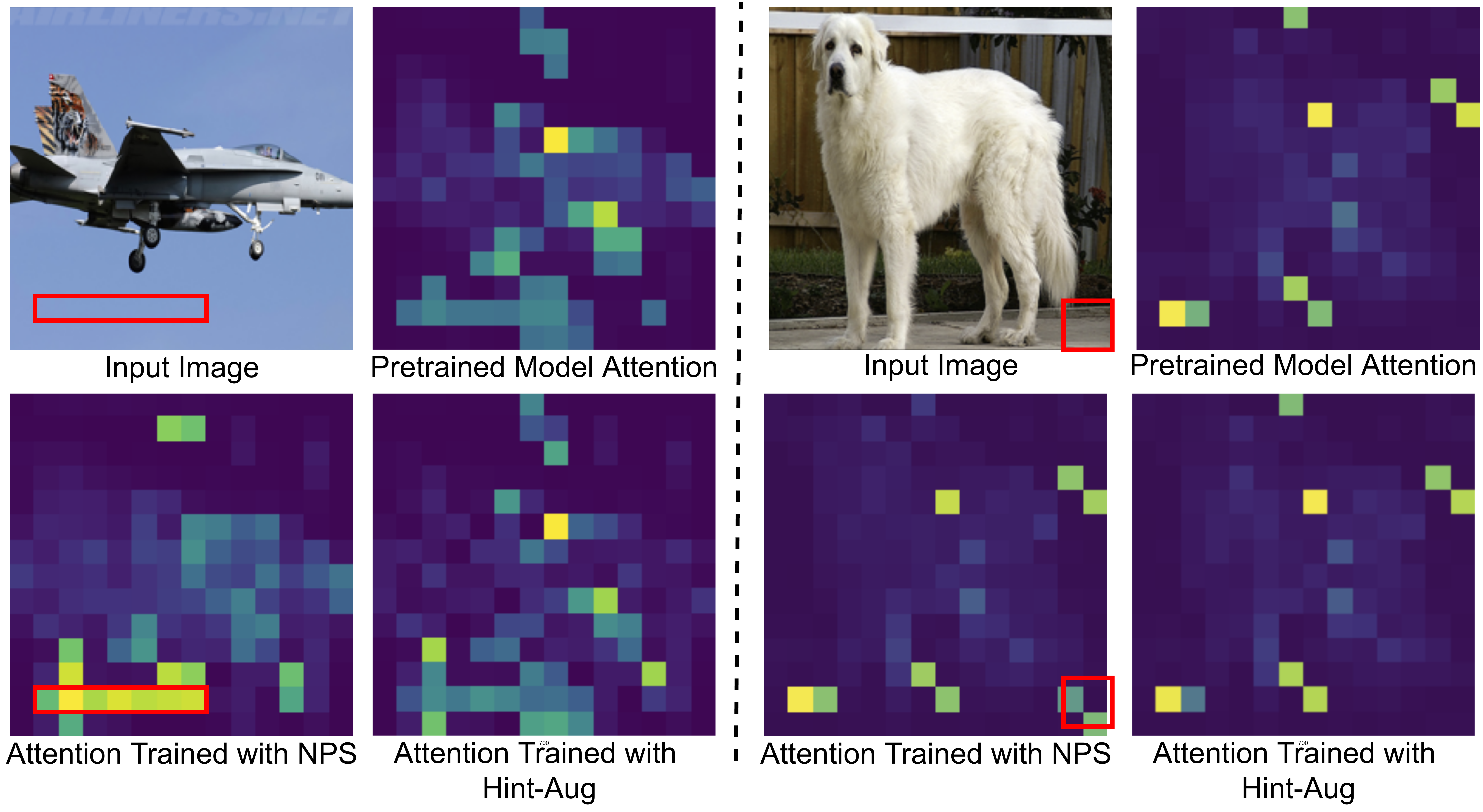}
    \caption{Visualization of attention score maps of images trained with different augmentation techniques.}
    \label{fig:all_attention}
\end{figure}

\begin{table}[t]
    \centering
        \caption{
        \cw{Impact of adversarial attack radius on the achievable accuracy of the Hint-Aug framework.}}
        \vspace{-1em}
    \begin{tabular}{c|ccccc}
    \toprule
       $\epsilon$  & 0.01 & 0.005 & 0.001 & 0.0002 & 0.0001\\
       \midrule
        Acc. (\%) & 70.01 & 70.83 & 71.04 & 71.01 & 70.15 \\ 
        \bottomrule
    \end{tabular}
    \vspace{-0.5em}
    \label{tab:aug_intensity}
\end{table}

\begin{table}[t]
    \centering
    \caption{Ablation on the number of selected patches to augment. }
    \vspace{-1em}
    \resizebox{\linewidth}{!}{
    \begin{tabular}{c|cccccc}
    \toprule
        \# patches & 1 & 2 & 3 & 8 & 32 & All\\
        \midrule
        Average Acc. & 65.42 & 65.44 & 65.35 &65.08  & 64.59 & 63.72 \\ 
        \bottomrule
    \end{tabular}
    }
    \vspace{-1em}
    \label{tab:num patch}
\end{table}

\vspace{-0.5em}
\subsubsection{Number of Patches to Augment}

Motivated by the promising accuracy-data efficiency trade-off achieved by Hint-Aug, an interesting question arises whether augmenting more than one patch for each image can further push forward the accuracy-efficiency trade-off. To answer this question, we conduct an ablation study on Hint-Aug with different numbers of augmented patches and report the average achieved accuracy when tuning with an 8-shot VPT~\cite{jia2022visual} across five datasets. Notably, augmenting all patches (i.e., column ``All" in Tab.~\ref{tab:num patch}) is equivalent to augmenting the whole image without considering the patch information. Our experiments show that augmenting one to three patches in each image leads to similar average accuracy (less than 0.1\% accuracy change). However, when augmenting more patches in the image, the average accuracy drops by $0.34\% \sim 1.70\%$ when augmenting more than 8 patches in each image. We suspect this is because only a few patches are prone to over-fitting in each image, as suggested in Fig.~\ref{fig:attn_vis}. Augmenting too many patches may ruin the attention-score map instead, leading to reduced accuracy.

\subsection{Visualization of Attention Score Maps}
To verify Hint-Aug's effectiveness in alleviating the over-fitting issue, we visualize the attention score maps of the pretrained FViT, FViT tuned by NPS~\cite{zhang2022neural}, and FViT tuned with our proposed Hint-Aug. As shown in Fig.~\ref{fig:all_attention}, we can observe that (1) after tuning with our proposed Hint-Aug, the over-fitted patches (marked in \textcolor{red}{red}) that are commonly observed in the attention score maps tuned by NPS~\cite{zhang2022neural} are successfully eliminated, and (2) the attention score map obtained from Hint-Aug features similar locations of high-attention score patches to those obtained from the pretrained FViT, indicating that Hint-Aug effectively alleviates the over-fitting issue. 


\subsection{Visualization of the Confusion Matrix}

\begin{wraptable}{r}{0.45\linewidth}
\centering
    \caption{The averaged confusion matrix value of the Cats and Dogs meta-group.}%
    \label{tab:cat&dog confusion}
    \resizebox{\linewidth}{!}{
    \begin{tabular}{c|cc} \toprule
  & Cats & Dogs \\ 
  \midrule
  Cats & 4.94 & 3.96\\
  Dogs & 3.96 & 5.72\\
  \bottomrule
  \end{tabular}
  }
\end{wraptable}

We visualize the confusion matrix using a 4-shot  Adapter~\cite{houlsby2019parameter} tuning setting on Pets~\cite{parkhi2012cats} to interpret the discovered class-wise similarity. We calculate the averaged confusion matrix value of the Cats and Dogs meta-group and visualize them in Tab.~\ref{tab:cat&dog confusion}. We observe that the FViT is much more confused in distinguishing between different classes within the Cat or Dog meta-group than distinguishing between the Cat and Dog meta-groups. This suggests that despite the simplicity of our strategy that uses the pre-softmax output, the generated confusion matrix can effectively identify the class pairs with easy-to-confuse features and thus provide correct guidance for CFI.


\vspace{-0.2em}
\section{Conclusion}
In this paper, we propose a framework called Hint-Aug, which is dedicated to boosting the few-shot parameter-efficient tuning accuracy of FViTs. Specifically, Hint-Aug features two enablers called AOD and CFI, aiming to alleviate the over-fitting issue and the lack of diverse data in few-shot tuning, respectively. Extensive experiments and ablation studies validate that Hint-Aug achieves a $0.04\%\sim32.91\%$ higher accuracy over SOTA data augmentation methods, opening up a new perspective towards more effectively tuning pretrained FViTs on downstream tasks in a realistic low-data scheme. 

\vspace{-0.2em}
\section*{Acknowledgement}
\zy{
The work was supported by the National Science Foundation (NSF) through the NSF CCF program (Award number: 2211815) and supported in part by CoCoSys, one of the seven centers in JUMP 2.0, a Semiconductor Research Corporation (SRC) program sponsored by DARPA.
}

{\small
\bibliographystyle{ieee_fullname}
\bibliography{egbib}
}

\end{document}


\title{Supplementary Materials of \\ Hint-Aug: Drawing Hints from Vision Foundation Models \\ towards Boosted Few-shot Parameter-Efficient ViT Tuning}

\author{First Author\\
Institution1\\
Institution1 address\\
{\tt\small firstauthor@i1.org}
\and
Second Author\\
Institution2\\
First line of institution2 address\\
{\tt\small secondauthor@i2.org}
}
\maketitle

\begin{table*}[]
    \centering
    \vspace{-1em}
    \caption{Benchmarking Hint-Aug with ViT-Base pretrained with MAE. Best results are marked in \textbf{bold}.}
    \vspace{-1em}
    \resizebox{0.6\linewidth}{!}{
    \begin{tabular}{c|c|c|ccccc}
    \toprule
       # Shot & Tuning Technique & Augment & Food & Pets & Cars & Flowers & Aircraft \\
        \midrule
       \multirow{9}{*}{8-shot}& \multirow{3}{*}{Adapter} & No-aug & 48.12 & 89.45 & 50.63 & 93.91 & 39.87\\
       & &  \cite{zhang2022neural} & 53.59 & 90.35 & 52.15 & 94.82 & 34.29\\
     & &  Proposed & \textbf{57.19} & \textbf{91.41} & \textbf{55.44} & \textbf{95.17} & \textbf{42.35}\\
       \cline{2-8}
       & \multirow{3}{*}{VPT} & No-aug & 49.13 & 92.12 & 48.97 & 89.64 & 38.79\\
        & &  \cite{zhang2022neural} & 53.12 & 90.71 & 47.74 & 94.36 & 36.12\\
       & &  Proposed & \textbf{55.85} & \textbf{92.48} & \textbf{49.29} & \textbf{94.83} & \textbf{40.15}\\
       \cline{2-8}
        & \multirow{3}{*}{LoRA} & No-aug & 46.39 & 88.31 & 47.68 & 93.75 & 39.62\\
        & &  \cite{zhang2022neural} & 53.16 & 89.75 & 55.81 & 95.70 & 38.61\\
        & &  Proposed & \textbf{55.11} & \textbf{90.22} & \textbf{57.12} & \textbf{96.01} & \textbf{41.10}\\
         \midrule
       \multirow{9}{*}{16-shot}& \multirow{3}{*}{Adapter} & No-aug & 58.41 & 91.09 & 73.56 & 95.86 &59.41 \\
       & &  \cite{zhang2022neural} & 63.05 & 91.58 & 75.03 & 96.77 & 54.97\\
     & &  Proposed & \textbf{65.01}& \textbf{92.45} & \textbf{77.54} & \textbf{97.48} & \textbf{62.42}\\
       \cline{2-8}
       & \multirow{3}{*}{VPT} & No-aug & 57.53 & 94.51 & 69.95 & 90.41 & 55.12\\
        & &  \cite{zhang2022neural} & 62.06 & 91.63 & 69.99 & 95.83 & 50.29\\
       & &  Proposed & \textbf{62.87} & \textbf{95.04} & \textbf{71.83} & \textbf{96.51} & \textbf{58.39}\\
       \cline{2-8}
        & \multirow{3}{*}{LoRA} & No-aug & 58.69 & 90.24 & 73.28 & 97.12 & 58.81 \\
        & &  \cite{zhang2022neural} & 62.37 & 91.93 & 77.74 & 97.65 & 56.62\\
        & &  Proposed & \textbf{63.10} & \textbf{92.38} & \textbf{78.81} & \textbf{98.10} & \textbf{59.49}\\
       \bottomrule
    \end{tabular}
    }
    \vspace{-0.5em}
    \label{tab:pretrain}
\end{table*}
\section{Overview and Outline}
\label{app:overview}
In this supplement, we provide more experiments and analysis as a complement to the main content, which are outlined below: 

\begin{itemize}
    \item We evaluate Hint-Aug's ability in boosting the FViTs' few-shot parameter-efficient tuning accuracy across \textbf{different pretraining strategies} in Supple.~\ref{app:pretrain}. 
    
    \item We evaluate Hint-Aug's ability in boosting the FViTs' accuracy under the less data-limited \textbf{transfer learning setting} in Supple.~\ref{app:transfer}. 
    
    \item We conduct an ablation study on Hint-Aug's sensitivity to $\lambda$ in Eq. 3 from the main paper in Supple.~\ref{app:lambda}.
\end{itemize}

\section{Hint-Aug's Achieved Accuracy Across Different Pretraining Strategies}
\label{app:pretrain}
The pretraining strategy is a key factor in fully exploiting the FViTs' potential~\cite{he2021masked,bao2021beit,chen2021empirical}. As mentioned in~\cite{he2021masked,chen2021empirical}, it plays an even more important role than the FViT models' architectures, given that we are now benchmarking under transformer-based models of similar architectures (e.g., applying the pretraining strategy as in~\cite{he2021masked} on ViT-Base can compensate for its accuracy gap with ViT-Huge, which has nine times more number of parameters). With this in mind, we aim to validate Hint-Aug's capability in boosting the FViTs' accuracy across models trained with different pretraining strategies. To do this, we validate the proposed Hint-Aug by applying it on top of the ViT-Base~\cite{dosovitskiy2020image} models pretrained with a widely used SOTA pretraining strategy, MAE~\cite{he2021masked}, while keeping all other experiment settings the same as those used in the experiments described in the main paper.

As shown in Table~\ref{tab:pretrain}, Hint-Aug can consistently boost the achievable accuracy over that of the baseline tuning methods by $0.31\% \sim 3.60\%$ and $0.45\% \sim 3.27\%$ on an 8-shot and 16-shot setting, respectively. This validates that Hint-Aug is a general method that can consistently improve the effectiveness of few-shot parameter-efficient tuning of FViTs.

\section{Hint-Aug's Ability in Boosting Transfer Learning Accuracy}
\label{app:transfer} 
We further validate Hint-Aug's achieved accuracy when we use more tuning data as in the conventional transfer learning scenario because it can evaluate Hint-Aug's potential as a general technique to improve parameter-efficient tuning accuracy under various scenarios. In this set of experiments, we tune the ImageNet pretrained ViT-Base~\cite{dosovitskiy2020image} on five datasets including Caltech-101~\cite{fei2004learning}, CIFAR-100~\cite{krizhevsky2009learning}, DTD~\cite{cimpoi2014describing}, Flowers~\cite{nilsback2006visual}, and Pets~\cite{parkhi2012cats}. We follow the same tuning setting in~\cite{zhang2022neural} and show our results in Table~\ref{tab:transfer}. We can see that Hint-Aug achieves as high as 2.47\%, 0.90\%, and 3.77\% accuracy improvement over the baseline method when tuning Adapter, VPT, and LoRA, respectively. This further validates that (1) Hint-Aug can generally boost the achievable accuracy under various settings in parameter-efficient tuning, and (2) the features generated in Hint-Aug are of high quality and can further improve the feature diversity even when more data (e.g., 50,000 tuning data in CIFAR-100 dataset) are provided.

\begin{table*}[]
    \centering
\caption{Benchmarking Hint-Aug with ViT-Base on transfer learning datasets. Best results are marked in \textbf{bold}.}
    \vspace{-1em}
    \resizebox{0.7\linewidth}{!}{
    \begin{tabular}{c|c|ccccc}
    \toprule
       Tuning Technique & Augment & Caltech-101 & CIFAR-100 & DTD & Flowers & Pets\\
       \midrule
       \multirow{3}{*}{Adapter} & No-aug & 90.10 & 69.22 & 68.04 & 98.61 & 89.88\\
        & \cite{zhang2022neural} & 88.58 & 64.79 & 70.32 & 98.68 & 90.61 \\
         & Proposed & \textbf{91.14} & \textbf{71.67 }&\textbf{ 70.64} & \textbf{99.01} & \textbf{90.86}\\
         \midrule
         \multirow{3}{*}{VPT} & No-aug & 90.83 & 76.76 & 65.78 & 98.03 & 88.27 \\
        & \cite{zhang2022neural} & 86.32 & 57.14 & 68.78 & 98.41 & 90.57\\
         & Proposed & \textbf{91.64} & \textbf{77.23} & \textbf{69.68} & \textbf{98.83} & \textbf{90.71} \\
         \midrule
         \multirow{3}{*}{LoRA} & No-aug & 91.37 & 67.12 & 69.40 & 98.55 & 90.38 \\
        & \cite{zhang2022neural} & 88.35 & 65.91 & 71.07 & 98.69 & 91.11\\
         & Proposed & \textbf{92.43} &\textbf{ 70.87} & \textbf{71.21} & \textbf{99.06} & \textbf{91.81}\\
         \bottomrule
    \end{tabular}
    }
    \vspace{-1em}
    \label{tab:transfer}
\end{table*}
\section{Ablation Study on $\lambda$'s Impact in Hint-Aug's Achievable Accuracy}
\label{app:lambda}
To validate Hint-Aug's ability when being applied to new tasks, we perform an ablation study regarding the impact of the only hyperparameter, $\lambda$ in Eq. 3, in Hint-Aug, under 8-shot Adapter tuning of ViT-Base~\cite{dosovitskiy2020image} on the Food~\cite{bossard2014food} dataset. As shown in Table~\ref{tab:lambda}, different values of $\lambda$ lead to marginal differences in the achievable accuracy (i.e., lower than 0.2\%). This validates that Hint-Aug is not sensitive to the hyperparameter being used, alleviating the potential burden of having to tune hyperparameters when applying Hint-Aug to a new task.

\begin{table}[]
    \centering
    \caption{Ablation study on $\lambda$'s impact.}
        \vspace{-1em}
    \begin{tabular}{c|ccc}
    \toprule
        $\lambda$ & 0.2 & 0.1 & 0.05 \\
        \midrule
        Accuracy (\%) & 70.88 & 71.04 & 70.85\\
         \bottomrule
    \end{tabular}
    \label{tab:lambda}
    \vspace{-1em}
\end{table}

{\small
\bibliographystyle{ieee_fullname}
\bibliography{egbib}
}